# Online Signature Recognition: A Biologically Inspired Feature Vector Splitting Approach

Marcos Faundez-Zanuy· Moises Diaz · Miguel Angel Ferrer

## Abstract

This research introduces an innovative approach to explore the cognitive and biologically inspired underpinnings of feature vector splitting for analyzing the significance of different attributes in e-security biometric signature recognition applications. Departing from traditional methods of concatenating features into an extended set, we employ multiple splitting strategies, aligning with cognitive principles, to preserve control over the relative importance of each feature subset. Our methodology is applied to three diverse databases (MCYT100, MCYT300, and SVC) using two classifiers (vector quantization and dynamic time warping with one and five training samples). Experimentation demonstrates that the fusion of pressure data with spatial coordinates ($x$ and $y$) consistently enhances performance. However, the inclusion of pen-tip angles in the same feature set yields mixed results, with performance improvements observed in select cases. This work delves into the cognitive aspects of feature fusion, shedding light on the cognitive relevance of feature vector splitting in e-security biometric applications.

## Introduction

Signature is one of the most widely used biometric traits for e-security, as it is based on "something you can do" [1, 2]. One of its main advantages is that the user can change their signature if it is compromised, which is unfortunately impossible with most other biometric traits such as the face, speech, and iris. Additionally, signature authentication has a long tradition of centuries in legal contracts, paintings, and other fields and can also play an essential role in health assessment [2, 3]. Several handwritten tasks can be used for signature recognition [4]. On-line biometric recognition can operate in two different ways:

Identification (1: $N$): The system compares the signature provided by the user with the $N$ models stored in the database of $N$ users. The model that best matches the input signature is used to identify the user.

Verification (1:1): A user provides their signature, and the system attempts to determine if they are a genuine or forged user by verifying their claimed identity. Some databases contain two types of forgeries, known as random and skilled. In the latter case, the forger is attempting to imitate the genuine signature, while in the former case, the forger is using their signature as a replacement for the genuine signature.

Online signature is probably the most popular behavioral trait for biometric recognition of people. Many individuals regularly sign documents, contracts, and other legal or financial papers, both in physical and digital formats. In addition, signature is the unique biometric trait that can be replaced (changed) if compromised. This is not possible or extremely difficult with other traits such as fingerprint face, in a broader field known as pattern recognition.

Figure 1 depicts the general pattern recognition scheme. It consists of:

The acquisition of a signal from the real world, usually by means of a digitizing tablet, smartphone, etc.

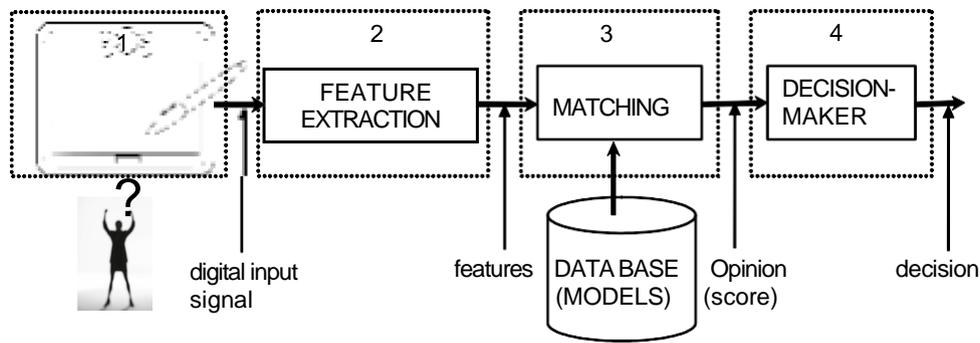

*Fig. 1 General pattern recognition system*

A feature extraction algorithm can extend the original set of features the digitizing device provides. An example is the popular delta and delta-delta parameters.These parameters are commonly used in various signal processing applications and are particularly useful for capturing the temporal changes or differences between consecutive values in a sequence. By adding delta and delta-delta parameters, the algorithm can provide additional information that characterizes the dynamic aspects of the signature, thereby enhancing the discriminative power and accuracy of the recognition system.

A matching algorithm that obtains a score (or a distance)measure that measures the matching degree between a pre-existing signature acquired during user enrollment and testing signature. Using the enrolment signatures, the computer works out a model usually stored in a data-base, card, etc.

Considering the score or distance value provided by thematching step, a decision is made, which is usually an "accept/reject the user" (verification application or 1:1 comparison) or the "identity" of the author of the testing signature (identification application or 1: $N$ comparison,where $N$ is the number of users contained in the data-base).

### Biometric Features Fusion

When dealing with extracted feature vectors, the goal is to combine the different measurements in order to improve theaccuracy of the decision. Several options exist, performing fusion at four different levels [5]:

1.Sensor level:

At this level, the digital input signal is the result of sensing the same biometric characteristic with two or more sensors. This first level is not usual in the online signatureacquisition, except for online and offline signature acquisition. An online signature is acquired through a digitizedtablet with paper on its top and an ink pen. Instead, an offline signature is obtained by scanning the signature produced on the sheet of paper. However, the goal of this combination is more focused on comparing online and offline systems than on the combination itself. To this end, simultaneous collection of offline and online datais needed. An example can be seen in the BiosecurID database [6]. In Galbally et al. [7], the authors combinedonline and enhanced offline data in separate automatic signature verifiers, which were combined at score level. A different strategy was proposed by Radhika and Gopika[8], where image-based and sequence-based techniques were applied to extract features of signatures to be combined in a single classifier.

Another type of work depends on the digitizers used. In some real applications, different tablets, smartphones, or PDAs can be used to collect and verify signatures. A proposal for sensor interoperability in signature verification can be seen in Alonso-Fernandez et al. [9]. The authors proposed an enrollment strategy using two tabletPC. In Tolosana et al. [10], a data preprocessing stage is highlighted to cope with the verification of signatures acquired in mobile and desktop scenarios.

## 2. Feature level:

This level implies obtaining an extended feature set by applying several algorithms on the basic feature set provided by the digitizing tablet, which normally is limited to five features: spatial coordinates ($x$, $y$), pressure ($p$), angles (azimuth, altitude). Among the most popular extended features are the first and second derivatives, delta and delta-delta parameters. This is a popular approach in speaker recognition [11, 12] and speech recognition [13] and signature recognition [14].

This combination strategy is usually done by concatenating the feature vectors extracted by each feature extractor. This yields an extended-size vector set.

Fifteen feature fusion proposals using spatial and pressure features were given by Parziale et al. [15]. They assessed the impact of each fusion with a standard and a stability-modulated DTW. In Diaz et al. [16, 17], a feature-level combination was carried out. The authors found good results by combining the position and pressure of a new set of features based on the movement of a robotic arm at the feature level. On the other hand, handcrafted and deep learning–based features were combined at the feature level [18]. The authors found that fusion eliminates the shortcomings of the corresponding group of features by complementing one another. Next, in Vorugunti et al.[19], a set of standard statistical-based features and deep representations from a convolutions auto-encoder were fused. The authors achieve competitive performance in few-shot learning scenarios.

While combining vectors is a popular fusion approach, it has some drawbacks, such as limited control over the contribution of each component to the final result and an increase in complexity for classifier design and data requirements. Therefore, opinion and decision levels are commonly used in state-of-the-art data fusion for pattern recognition. This allows for greater control over the fusion process and a better understanding of the contribution of each component to the final result.

## 3. Opinion level:

Fusing different biometric classifiers to achieve better performance is also known as confidence level fusion. This approach combines the scores provided by each matcher and the distance or similarity measure between the input features and the models stored in the database. It is possible to combine several classifiers working with the same biometric characteristic or different ones altogether, known as unimodal and multimodal fusion.

Before the fusion process can begin, normalization must be performed. For example, suppose one classifier produces similarity measures within the [0, 1] range and another produces distance measures within the [0, 100] range. In that case, two normalization steps must be taken:

-Converting the similarity measures into distance measures or vice versa.

-Standardizing the location and scale parameters of the similarity scores from each classifier.

After normalization, several combination schemes can be applied [20].

The combination strategies can be classified into three main groups:

Fixed rules: In this case, all the classifiers have the same relevance. An example is the sum of the out-puts of the classifiers. In Okawa [21], a simple sum of two outputs scores was used for the final score value in online signature verification. Independent and dependent warping strategies in DTW obtained the two individual scores. Competitive performances were obtained with three public online signature datasets: SVC2004 Task1/Task2 and MCYT-100.

Trained rules: Differently, some classifiers should have more relevance to the final result. It is achieved through weighting factors computed using a training sequence. The most straightforward procedure when dealing with two classifiers with their respective opinions $o_1$ and $o_2$ is obtaining a combined opinion:

$o = \alpha o_1 + (1 - \alpha)o_2$, where $\alpha$ is the weighting factor optimized by finding the highest accuracy for $\alpha \in [0, 1]$

An example of the combination of verification scores through a weighted mean can be seen in Fischer and Plamondon [22]. They combined a well- established verification using dynamic time warping with lognormal-based features with a string edit distance strategy. The performance was quite competitive compared to several benchmark datasets.

In a study conducted by Diaz et al. [17], they introduced a novel approach that combined stand- ard features and robotic features at the score level, yielding the most optimal results as reported in their research article. Furthermore, their findings demonstrated the superior performance achieved through the feature fusion strategy employed in this modality.

Adaptive rules: The relevance of each classifier depends on the instant time. It is interesting for variable environments, and it is a generalization of the previous rule:

$o = \alpha(t)o_1 + (1 - \alpha(t))o_2$, where $\alpha(t)$ is the weighting factor, which is variable in certain conditions.

The most popular combinations are weighted sum, weighted product, and decision trees (based on if–then–else sentences).

4. Decision level:

   The fusion of multiple classifiers is required at this level. Each classifier generates a decision, which is either an accepted/rejected decision for verification applications or the identity of the person or a ranked list with the most probable person on top for identification systems. In the latter case, the Borda count method [23] can be used to combine the outputs of classifiers. This method circum- vents the mandatory score normalization for the opinion fusion level and assigns a score equal to the number of classes ranked below the given category.

   However, one issue with decision-level fusion is the pos-sibility of ties. At least three classifiers are necessary for verification applications, as two classifiers must agree to avoid ties. For identification scenarios, the number of classifiers must be higher than the number of classes, which is typically impractical. Consequently, this com- bination level is typically used for verification scenarios. A significant combination scheme at the decision level is the serial and parallel combination, also known as the "AND" and "OR" combinations. The "AND" combination enhances the false acceptance rate (FAR), while the "OR" combination improves the false rejection rate (FRR).

   By simultaneously combining serial and parallel systems, it is possible to enhance both rates [24]. Existing proposals for online signature recognition that extend feature sets are based on the feature level, where new features are concatenated to form an extended feature set. This paper will investigate various combination strategies at the opinion level to better understand each feature's contribution to the final accuracy. More specifi- cally, our proposal allows us to determine the relevance of pressure and pen-tip angle information in e-security biometric application.

**Experimental Setup**
*Database*

We have used two different databases in our experimental section:

   *MCYT database*: We utilized the MCYT database [25] in its entirety, which comprises two sets of signatures: one using a software animation viewer of the signature to be forged. For this study, we utilized a final set of 16,000 sig- natures (8000 genuine signatures and 8000 skilled forgeries), approximately 10% of the size of the MCYT database.

   It is worth highlighting the signatures in the SVC data- base are primarily in English or Chinese, and no genuine signatures were used. Instead, contributors were advised to create and practice a new signature before the acquisition sessions.

**Feature Set Extension and Normalization**

Starting from the basic set of features $f = x, y, p, az, al,$ where and $y$ are the spatial coordinates, $p$ the pen pressure on the tablet, and $az$ and $al$ the azimuth and zenith angles, provided by the digitizing tablet we extended it by workingout the delta and delta-delta parameters.

Delta ($\dot{f_i}$) and delta-delta ($\ddot{f_i}$) features are the first and second derivative, respectively, for each basic feature $f_i$, where $i \in [1, 5]$. The delta parameters are obtained in the following way with 330 users (MCYT330) and another with a subset of [27]:

$$\dot{f_i} = \frac{\sum_{k=-M}^{M} k \cdot f[k]}{\sum_{k=-M}^{M} k^2} \tag{1}$$

The delta of a given feature $f$ can be approximated through the least-squares method as the local slope within a region surrounding the current sample $f[k]$. This region encompasses $M$ samples preceding and succeeding the current one, resulting in $2M + 1$ sample. The delta window length parameter determines the size of this region and is defined from $-M$ to $M$. To calculate the delta, an odd integer greater than or equal to three must be specified as the window length.

Delta-delta is obtained by applying two consecutive times the delta Eq. (1).

Features are normalized through a z-score using the following equation, where each feature fi is subtracted by its mean and divided by its standard deviation.

$$\hat{f_i} = \frac{f_i - \bar{f_i}}{\text{std}(f_i)} \tag{2}$$

**Modifying the Feature Vector Length**

The signature acquired by digitizing tablet at a sampling rate of 100 samples per second consists of a set of $L$ samples, where each sample contains five different values $x, y, p, az, al,$ provided by the digitizing tablet plus the delta and delta-delta parameters. Without loss of generality, we will describe the splitting process to obtain a couple of different feature sets from the original five-dimensional feature set.

There are several basic strategies to modify the feature vector length and the sequence of feature vectors, as shown in Fig. 2:

1. Use the whole set of features $(x, y, p, a, z)$ to calculate one model per user (model1 in Fig. 2, obtained from a sequence of L samples, each sample 5 features).

2. Build two subvectors by splitting the features and generate one model for each subset (model2a and model2b in Fig. 2, obtained from a sequence of L samples being each sample 2 features for model2a and three features for model2b. Obviously, different feature separations are possible).

3. Designing the feature vectors by concatenating the features of $S$ consecutive sampling points (with)out overlapping. In this example (model2, Fig. 2), $S = 2$ but of course different values are possible).

4. Split the feature set into several sections such as initial, middle, and ending section. In this approach, the feature vector dimension is not modified, but the whole sequence of feature vectors (L) is split into sections. This approach has been applied in a vector quantization algorithm known as multi-section vector quantization [28].

One potential application or relationship between feature vector splitting and cognitive computation lies in the field of cognitive pattern recognition or cognitive modeling. Cognitive computation aims to develop computational models that mimic human cognitive processes and abilities, such as perception, attention, memory, and decision-making.

Feature vector splitting can be used as a technique to simulate how humans process and analyze information in a hierarchical or segmented manner. Human cognition often involves breaking down complex stimuli or patterns into smaller, more manageable components. By splitting the feature vector into subsets or segments, cognitive models can mimic this hierarchical processing and capture the sequential or parallel nature of human cognitive operations.

In cognitive pattern recognition tasks, such as object recognition or scene understanding, feature vector splitting can help replicate the cognitive processes of selective attention and feature integration. By splitting the feature vector into subsets that correspond to different ways to parametrize a signature, such as geometrical, texture, and shape, cognitive models can focus attention on specific subsets and integrate the information from different subsets to form a holistic representation of the pattern or stimulus.

Furthermore, feature vector splitting can be employed in cognitive computation to investigate how humans prioritize and weight different features during pattern recognition tasks. By selectively splitting the feature vector based on specific feature properties or relevance criteria, cognitive models can simulate human feature selection strategies and explore the impact of feature weighting on pattern recognition performance.

Our motivation for this approach stems from the following cognitive insights:

1. Hierarchical Processing: Recent research in cognitive neuroscience [29] suggests that the brain processes interpersonal verbal communication hierarchically. They approach this issue by proposing three levels of neurocognitive processes are required. Such a division of the information is also proposed in our feature vector splitting, which aligns with this hierarchical processing by preserving control over individual feature subsets.

2. Selective Attention: Cognitive neuroscience studies [30] emphasize the role of selective attention in visual processing. Our approach allows us to selectively emphasize or de-emphasize specific features, simulating the cognitive process of selective attention for biometric recognition.

3. Perceptual Grouping: The Gestalt principles of perceptual grouping have been instrumental in understanding how humans organize visual information. In Prieto et al. [31], the authors conclude with evidences about improvements in visual working memory when part of the information is grouped through perceptual grouping cues. Our feature vector splitting strategy enables us to explore how grouping specific features affects biometric recognition performance.

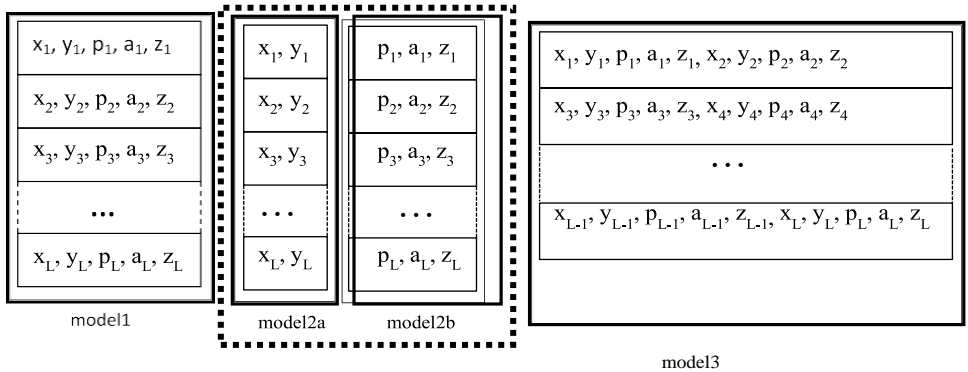

**Fig. 2** *Strategies to split the feature vectors by features (model2a, model2b) and to increase the dimensionality of the vectors (model3)*

Overall, the application of feature vector splitting in cognitive computation allows for the development and exploration of computational models that capture the hierarchical, selective, and integrative aspects of human cognition. By incorporating this technique, researchers can gain insights into human cognitive processes and potentially improve the performance of cognitive systems for various tasks, such as perception, decision-making, and problem-solving.

A simultaneous combination of strategies 2 and 3 is also an option. However, our experimental results revealed poor performance when using strategy three (model3 in Fig. 2), prompting us to exclude combinations of 2 and 3. Probably this strategy yields worse outcomes due to the curse of dimensionality: the higher the vector dimensions, the more challenging to classify vectors.

Considering the different fusion levels described in the "Introduction" section in the first approach depicted in Fig. 2 (model1), we are in the second situation: the digitizing tablet detects the two spatial coordinates, the pressure, and a couple of angles, and we concatenate them to obtain a concatenated feature vector of five dimensions. One of the main problems of this approach is the little control over the contribution of each component vector on the final result, and the augmented feature space can imply a more rugged classifier design, the need for more training and testing data, etc. [24]. Although it is a common belief that the earlier the combination is done, the better result is achieved, state-of-the-art data fusion relies mainly on the opinion level.

When using the second approach depicted in Fig. 2 (model2a + model2b), we are shifting part of the feature level fusion to opinion fusion, as the final result is obtained by combining the scores provided by the combination of two different classifiers.

In this paper, we have defined four different scenarios when splitting the feature vector into two sets, where each set feeds one classifier:

TEST 1 $[\underbrace{x, y, \dot{x}, \dot{y}, \ddot{x}, \ddot{y}}_{set1}]$, $[\underbrace{p, \dot{p}}_{set2}]$: no pressure neither angle in set1, the pressure goes into a second classifier.

TEST 2 $[\underbrace{x, y, \dot{x}, \dot{y}, \ddot{x}, \ddot{y}}_{set1}]$, $[\underbrace{p, az, al, \dot{p}, \dot{az}, \dot{al}}_{set2}]$: no pressure neither angles in set1, pressure and angles in the second classifier.

TEST 3 $[\underbrace{x, y, p, \dot{x}, \dot{y}, \dot{p}, \ddot{x}, \ddot{y}}_{set1}]$, $[\underbrace{az, al, \dot{az}, \dot{al}}_{set2}]$: set1 includes pressure, and angles in a separate classifier.

TEST4 $[\underbrace{x, y, p, \dot{x}, \dot{y}, \dot{p}, \ddot{x}, \ddot{y}}_{set1}]$, $[\underbrace{p, az, al, \dot{p}, \dot{az}, \dot{al}}_{set2}]$: pressure is included in set1 and set2, angles in the second set.

**Classification Algorithms**

We have used two classification algorithms:

Vector quantization (VQ): One possibility for biometric recognition consists of modeling each user by its own codebook [32], where the Linde-Buzo-Gray (LBG) k-means algorithm generates the codebook [33], and the number of users inside the database is N. We have used the LBG algorithm implementation available in VOICEBOX: Speech Processing Toolbox for MATLAB.

Several signatures, usually acquired in the same session (same day), are used to train the codebook. Then during the test, the process is the following, given a test signature acquired in a different acquisition session (day) than the training one:

a) Identification (1:N): the test signature is quantized with the whole set of codebooks, and one quantization distortion (distance measure) is obtained for each codebook (user). The codebook that provides the smallest distance reveals the identity of the user.

b) Verification (1:1): the test signature is quantized with the codebook belonging to the claimed identity. If this distance is smaller than the decision threshold, the user is accepted (considered a genuine user). Otherwise, it is rejected (considered an impostor).

As far as the feature set has been split into two different sets, the process is the following:

1. Using the training signatures, one codebook is obtained for each set (CB1, CB2; named model2a and model 2b in Fig. 2).

2. Each testing signature is split into two sets (set1, set2), and each part is quantized with its corresponding codebook providing a quantization distortion d(CB1, set1) and d(CB2,set2).

3. The two quantization distortions obtained for each user and testing signature are combined by a trained rule as described in Sect. 1:
combined = $\alpha \cdot$ d (CB1, set1) + (1 - $\alpha$) $\cdot$ d(CB2, set2)

Dynamic time warping (DTW): Feature matching of the testing sample with the models is performed using DTW. It is a popular template-matching algorithm well suited to cope with random variations due to intra-user variability [34]. DTW applies a dynamic programming strategy to produce an elastic measure of the distance between two samples even if they differ in length. We proceed in an analogous way to the VQ algorithm. First, we compute a couple of DTW distances, one with model2a (first part of the feature set) and another one with model 2b (last part of the feature set). Then, we have analyzed the DTW algorithm under two training conditions: DTW1 (1 training signature per user) and DTW5 (5 training signatures per user). The DTW distance is computed between each training signature and the testing sample in the last case. From the set of five distances obtained, the minimum distance is selected.

Deep learning techniques could also be employed. They offer high classification rates and eliminate the need for feature vector extension and normalization. However, we have chosen to exclude these algorithms from the scope of this paper for the following reasons:

a) Limited data availability: Deep learning algorithms typically require a large amount of labeled data to achieve optimal performance. In the field of biometric signature recognition, acquiring a large number of training signatures per user can be time-consuming, cumbersome for the users and unpractical in real-world scenarios. On the other hand, classical machine learning algorithms such as VQ and DTW require less data and can still yield reliable results with a smaller dataset.

b) Interpretability: The goal of our paper is to shed light on the contribution of specific features obtained by digitizing tablet. For instance, can we discard pressure or angles? The use of deep learning techniques, although it can provide good classification accuracies, is sometimes like a black box where you obtain the result, but interpretability of selected features and their relative importance is overlooked.

c) Computational efficiency: Deep learning models, especially those with numerous layers and parameters, can be computationally intensive and require significant computational resources. In contrast, the VQ and DTW algorithms used in this paper offer greater computationally efficient.

**Experimental Results**

Tables 1, 2, 3, and 4 present the experimental results using VQ with different bits ranging from 4 to 8 (this impliesa separate codebook or model size) and three databases: MCYT330, MCYT100, and SVC. Instead, Table 5 shows the results with all tests and DTW signature verifier. The best results are highlighted in color in these tables.

We have utilized two distinct automatic signature verifiers: DTW and VQ. While DTW observes the temporal sequence of features, VQ models signatures with multiple codebooks, where the number of bits defines the size ofthe codebook. The LBG algorithm is used to construct the codebook. It begins by generating a codebook with a singlevector (bit = 0; 2^0 = 1) and then doubles its size in the next iteration. To accomplish this, one vector is taken from the previous iteration, and a new vector is produced by addinga random disturbance to the first vector. The positions of thecentroids are iteratively readjusted to ensure they best represent the training sequence. When convergence is achieved, We have obtained the following experimental measures:

Identification: The identification rate (IDR). It is calculated by dividing the number of correctly identified individuals by the total number of individuals in the population.

Verification: Detection cost function (DCF). The DCFis a metric that measures the cost of making errors in the authentication process. It is typically expressed as a weighted sum of the false acceptance rate (FAR) and the false rejection rate (FRR), where the weights represent the relative cost of each type of error. The minimum of the detection cost function (minDCF) refers to the pointat which the overall cost of the system is minimized. This point represents the optimal operating point for thesystem, where the trade-off between FAR and FRR is balanced to minimize the overall cost. DCFr and DCFs are, respectively, the minimum value of the detection cost function (minDCF) for random (DCFr) and skilled (DCFs) forgeries. Mathematically, DCF is defined in Martin et al. [35] as $DCF = C_{miss} \times P_{miss} \times P_{true} + P_{fa} \times P_{false}$, where $C_{miss}$ is the cost of a miss (rejection), $C_{fa}$ is the cost of a false alarm (acceptance), $P_{true}$ is thea priori probability of the target, and $P_{false} = 1 - P_{true}$. In our setup, we have configured $C_{miss} = C_{fa} = 1$.

*Table 1  Results with Test 1, different databases and VQ-based automatic signature verifier $\alpha_{OPT}$ denotes the optimal alpha value for considering both set of features at score level. $\alpha = 0$ and $i = 1$ mean that only the second or first set of features, respectively, was evaluated. Green color highlights the optimal VQ configuration for random forgeries. Yellow color highlights the optimal VQ configuration for skilled forgeries. This is always obtained for codebook sizes greater or equal to six bits.*

| | | b=4 | | | b=5 | | | b=6 | | | b=7 | | | b=8 | | |
|---|---|---|---|---|---|---|---|---|---|---|---|---|---|---|---|---|---|
| | | IDR | DCFr | DCFs | IDR | DCFr | DCFs | IDR | DCFr | DCFs | IDR | DCFr | DCFs | IDR | DCFr | DCFs |
| MCYT | $\alpha_{OPT}$ | 98 | 4.31 | 13.66 | 99 | 3.90 | 10.44 | 98.6 | 2.91 | 8.7 | 98 | 2.84 | 7.86 | 97.4 | 3.35 | 8.48 |
| VQ1_100 | $\alpha=0$ | 26.2 | 28.52 | 30.5 | 25.8 | 28.65 | 28.86 | 22.4 | 29.33 | 29.24 | 18.2 | 30.66 | 29.78 | 16.8 | 30.74 | 30.60 |
| | $\alpha=1$ | 97 | 5.73 | 18.14 | 98.4 | 4.10 | 13.12 | 98.6 | 3.40 | 11.18 | 97.8 | 3.03 | 9.88 | 97.4 | 3.44 | 9.54 |
| MCYT | $\alpha_{OPT}$ | 100 | 2.25 | 9.58 | 100 | 1.42 | 6.74 | 100 | 0.97 | 5.32 | 100 | 0.82 | 4.62 | 100 | 0.90 | 4.18 |
| VQ5_100 | $\alpha=0$ | 42 | 23.64 | 26.38 | 48.2 | 21.34 | 23.92 | 44.4 | 20.78 | 22.26 | 42 | 21.20 | 21.8 | 35.2 | 21.69 | 21.64 |
| | $\alpha=1$ | 99.2 | 4.1 | 14.4 | 100 | 2.14 | 9.96 | 99.8 | 1.48 | 7.66 | 100 | 1.27 | 6.62 | 100 | 1.22 | 5.66 |
| MCYT | $\alpha_{OPT}$ | 92.06 | 5.12 | 13.50 | 94.06 | 4.06 | 11.33 | 94.91 | 3.59 | 10.38 | 94.73 | 3.70 | 9.67 | 93.27 | 3.79 | 9.75 |
| VQ1_330 | $\alpha=0$ | 17.15 | 27.56 | 30.51 | 16.67 | 27.82 | 29.72 | 13.88 | 27.96 | 29.18 | 11.39 | 29.31 | 30.05 | 9.45 | 29.90 | 30.83 |
| | $\alpha=1$ | 90.55 | 6.34 | 16.48 | 93.39 | 4.75 | 13.36 | 94.67 | 3.94 | 11.97 | 94.42 | 3.87 | 10.96 | 92.85 | 4.06 | 10.41 |
| MCYT | $\alpha_{OPT}$ | 98.79 | 2.27 | 8.91 | 99.09 | 1.46 | 6.71 | 99.15 | 1.21 | 5.07 | 98.97 | 1.15 | 4.39 | 98.91 | 1.04 | 3.96 |
| VQ5_330 | $\alpha=0$ | 31.81 | 22.54 | 27.45 | 35.33 | 20.49 | 24.74 | 33.27 | 20.17 | 23.22 | 29.39 | 20.33 | 22.34 | 24.73 | 21.16 | 22.22 |
| | $\alpha=1$ | 98.12 | 3.71 | 12.59 | 98.91 | 2.28 | 9.25 | 99.09 | 1.72 | 7.00 | 98.91 | 1.36 | 5.92 | 98.85 | 1.27 | 5.24 |
| SVC | $\alpha_{OPT}$ | 99 | 0.92 | 24.19 | 99 | 0.40 | 21.44 | 99.5 | 0.29 | 20.31 | 99.5 | 0.41 | 18.5 | 99.5 | 0.77 | 18.5 |
| VQ1 | $\alpha=0$ | 17.5 | 33.60 | 40.63 | 20.5 | 33.15 | 39.75 | 15.5 | 35.65 | 41.63 | 11 | 37.06 | 41.94 | 10 | 37.94 | 42.63 |
| | $\alpha=1$ | 99 | 1.13 | 24.75 | 99 | 0.77 | 21.75 | 99.5 | 0.32 | 20.56 | 99.5 | 0.42 | 18.75 | 99.5 | 0.78 | 18.69 |
| SVC | $\alpha_{OPT}$ | 100 | 0.32 | 21.81 | 100 | 0.04 | 16.38 | 100 | 0 | 12.06 | 100 | 0 | 10.56 | 100 | 0.03 | 8.56 |
| VQ5 | $\alpha=0$ | 35 | 22.76 | 32.44 | 42.5 | 21.85 | 30.88 | 45 | 21.90 | 29.38 | 37 | 22.93 | 30.5 | 30.5 | 24.77 | 31 |
| | $\alpha=1$ | 100 | 0.74 | 22.75 | 100 | 0.15 | 17.13 | 100 | 0.01 | 12.5 | 100 | 0 | 10.75 | 100 | 0.03 | 8.56 |

*Table 2  Results with Test 2, different databases and VQ-based automatic signature verifier α OPT denotes the optimal alpha value for considering both set of features at score level. α =0 and α= 1 mean that only the second or first set of features, respectively, was evaluated. Green color highlights the optimal VQ configuration for random forgeries. Yellow color highlights the optimal VQ configuration for skilled forgeries. This is always obtained for codebook sizes greater or equal to six bits*

| | | b=4 | | | b=5 | | | b=6 | | | b=7 | | | b=8 | | |
|---|---|---|---|---|---|---|---|---|---|---|---|---|---|---|---|---|
| | | IDR | DCFr | DCFs | IDR | DCFr | DCFs | IDR | DCFr | DCFs | IDR | DCFr | DCFs | IDR | DCFr | DCFs |
| MCYT VQ1_100 | $\alpha_{OPT}$ | 96.8 | 5.14 | 14.96 | 98.8 | 3.13 | 11.3 | 98.6 | 2.83 | 9.74 | 98.4 | **2.80** | 8.54 | 97.4 | 3.24 | **8.34** |
| | $\alpha=0$ | 17.2 | 26.99 | 30.4 | 18.2 | 26.10 | 29.1 | 20.0 | 27.05 | 29.5 | 17.8 | 27.68 | 29.44 | 15.6 | 30.45 | 31.24 |
| | $\alpha=1$ | 95.8 | 5.93 | 17.9 | 98.0 | 4.10 | 12.9 | 98.2 | 3.43 | 11.34 | 98.0 | 3.00 | 9.94 | 97.4 | 3.44 | 9.54 |
| MCYT VQ5_100 | $\alpha_{OPT}$ | 100 | 2.61 | 11.08 | 100 | 1.35 | 7.48 | 100 | 0.93 | 5.56 | 100 | 0.86 | 4.72 | 100 | **0.74** | **3.92** |
| | $\alpha=0$ | 39.8 | 20.08 | 25 | 47 | 17.28 | 22.46 | 49.2 | 16.98 | 21.98 | 49.8 | 15.74 | 20.24 | 52.4 | 15.94 | 20.52 |
| | $\alpha=1$ | 99.6 | 3.77 | 13.96 | 100 | 2.04 | 10.10 | 100 | 1.34 | 7.94 | 100 | 1.30 | 6.30 | 100 | 1.21 | 5.64 |
| MCYT VQ1_330 | $\alpha_{OPT}$ | 90.42 | 5.85 | 14.38 | 94.24 | 4.31 | 11.84 | 95.09 | **3.58** | 10.55 | 94.79 | 3.72 | **9.68** | 93.21 | 3.94 | 9.82 |
| | $\alpha=0$ | 13.76 | 27.35 | 32.12 | 15.33 | 27.00 | 31.40 | 15.64 | 26.42 | 30.64 | 14.55 | 27.48 | 30.96 | 13.21 | 29.47 | 32.06 |
| | $\alpha=1$ | 89.39 | 6.55 | 16.32 | 93.52 | 4.82 | 13.53 | 94.48 | 4.05 | 11.88 | 94.48 | 3.80 | 10.74 | 92.67 | 4.03 | 10.37 |
| MCYT VQ5_330 | $\alpha_{OPT}$ | 98.55 | 2.87 | 10.19 | 99.09 | 1.82 | 7.30 | 99.15 | 1.27 | 5.59 | 99.09 | 1.45 | 4.55 | 99.03 | **1.12** | **3.87** |
| | $\alpha=0$ | 31.27 | 19.20 | 26.02 | 36.18 | 17.45 | 24.70 | 39.09 | 16.91 | 24.28 | 42.18 | 16.64 | 23.72 | 41.82 | 16.12 | 23.55 |
| | $\alpha=1$ | 97.70 | 3.79 | 12.81 | 98.67 | 2.14 | 9.49 | 98.91 | 1.60 | 7.19 | 98.97 | 1.32 | 6.01 | 98.85 | 1.24 | 5.16 |
| SVC VQ1 | $\alpha_{OPT}$ | 99.5 | 0.77 | 20.75 | 99.5 | 0.37 | 17.73 | 99.5 | **0.32** | 15.94 | 99.5 | 0.44 | **15.06** | 99.5 | 0.72 | 16.94 |
| | $\alpha=0$ | 61 | 16.22 | 29.88 | 63.5 | 16.58 | 28.75 | 63.5 | 16.34 | 27.44 | 60.5 | 16.63 | 27.44 | 53.5 | 18.42 | 28.5 |
| | $\alpha=1$ | 99 | 1.24 | 27.44 | 99 | 0.51 | 22.13 | 99.5 | 0.32 | 21.19 | 99.5 | 0.44 | 19.31 | 99.5 | 0.77 | 18.69 |
| SVC VQ5 | $\alpha_{OPT}$ | 100 | 0.01 | 12.56 | 100 | 0.01 | 10.13 | 100 | 0.01 | 8.13 | 100 | **0** | 6.69 | 100 | 0.01 | **5.69** |
| | $\alpha=0$ | 88 | 7.98 | 20.31 | 89.5 | 8.13 | 18.06 | 92 | 6.67 | 15.63 | 91 | 6.73 | 15.44 | 93 | 6.56 | 15.25 |
| | $\alpha=1$ | 100 | 0.64 | 21.19 | 100 | 0.08 | 15.81 | 100 | 0.01 | 12.56 | 100 | 0 | 10.44 | 100 | 0.01 | 9.19 |

*Table 3  Results with Test 3, different databases and VQ-based automatic signature verifier $\alpha_{OPT}$ denotes the optimal alpha value for considering both set of features at score level. α = 0 and α = 1 mean that only the second or first set of features, respectively, was evaluated. Green color highlights the optimal VQ configuration for random forgeries. Yellow color highlights the optimal VQ configuration for skilled forgeries. This is always obtained for codebook sizes greater or equal to six bits.*

| | | b=4 | | | b=5 | | | b=6 | | | b=7 | | | b=8 | | |
|---|---|---|---|---|---|---|---|---|---|---|---|---|---|---|---|---|
| | | IDR | DCFr | DCFs | IDR | DCFr | DCFs | IDR | DCFr | DCFs | IDR | DCFr | DCFs | IDR | DCFr | DCFs |
| MCYT VQ1_100 | $\alpha_{OPT}$ | 97.4 | 3.65 | 13.38 | 99 | 1.98 | 9.44 | 99 | **1.49** | 8.34 | 98.8 | 1.93 | 7.12 | 98 | 2.46 | **6.98** |
| | $\alpha=0$ | 6 | 40.23 | 37.54 | 5 | 39.78 | 36.72 | 6 | 39.36 | 35.64 | 5 | 40.81 | 36.06 | 3.4 | 42.40 | 37.10 |
| | $\alpha=1$ | 96.8 | 3.84 | 13.46 | 99.0 | 2.00 | 9.66 | 99.0 | 1.50 | 8.48 | 98.6 | 1.93 | 7.54 | 98.0 | 2.50 | 7.00 |
| MCYT VQ5_100 | $\alpha_{OPT}$ | 99.8 | 2.09 | 10.5 | 100 | 0.86 | 7.44 | 100 | 0.68 | 5.44 | 100 | 0.49 | 4.36 | 100 | **0.40** | **3.64** |
| | $\alpha=0$ | 12.4 | 33.59 | 31.78 | 14.8 | 32.36 | 30.82 | 16.2 | 32.75 | 31.40 | 13.8 | 32.39 | 30.30 | 14 | 32.11 | 30.36 |
| | $\alpha=1$ | 99.6 | 2.30 | 11.78 | 100 | 0.86 | 7.9 | 100 | 0.69 | 5.56 | 100 | 0.50 | 4.48 | 100 | 0.40 | 3.78 |
| MCYT VQ1_330 | $\alpha_{OPT}$ | 92.79 | 4.26 | 13.09 | 95.94 | 2.88 | 10.39 | 96.48 | **2.38** | 8.89 | 96.67 | 2.48 | 8.29 | 96.06 | 2.76 | **8.27** |
| | $\alpha=0$ | 4.55 | 39.00 | 38.13 | 4.24 | 39.03 | 37.64 | 4.24 | 39.10 | 37.41 | 4.24 | 40.27 | 37.52 | 3.58 | 41.88 | 38.21 |
| | $\alpha=1$ | 92.61 | 4.35 | 13.21 | 95.76 | 2.90 | 10.46 | 96.48 | 2.40 | 8.90 | 96.67 | 2.48 | 8.38 | 96.06 | 2.76 | 8.27 |
| MCYT VQ5_330 | $\alpha_{OPT}$ | 99.03 | 2.14 | 9.84 | 99.21 | 1.11 | 6.8 | 99.45 | 0.83 | 4.83 | 99.45 | 0.71 | 4.13 | 99.45 | **0.67** | **3.53** |
| | $\alpha=0$ | 9.82 | 31.48 | 32.15 | 10.42 | 31.47 | 32.37 | 11.94 | 30.92 | 32.30 | 11.45 | 30.62 | 31.71 | 11.58 | 30.92 | 32.02 |
| | $\alpha=1$ | 99.03 | 2.17 | 10.22 | 99.21 | 1.15 | 7.01 | 99.45 | 0.83 | 4.95 | 99.39 | 0.71 | 4.19 | 99.39 | 0.67 | 3.55 |
| SVC VQ1 | $\alpha_{OPT}$ | 98.5 | 1.16 | 18.81 | 99.5 | 0.91 | 16.44 | 99.5 | **0.79** | 14.94 | 99.5 | 1.29 | 14.31 | 99.5 | 1.57 | **13.81** |
| | $\alpha=0$ | 46.5 | 22.99 | 31.63 | 48 | 23.42 | 31.31 | 48.5 | 22.87 | 31.19 | 46 | 24.16 | 32.38 | 37.5 | 25.68 | 33.31 |
| | $\alpha=1$ | 98.5 | 1.25 | 22 | 99.5 | 0.91 | 19.44 | 99.5 | 0.79 | 17.13 | 99.5 | 1.29 | 15.94 | 99.5 | 1.57 | 14.81 |
| SVC VQ5 | $\alpha_{OPT}$ | 100 | 0.31 | 13.44 | 100 | 0 | 9.56 | 100 | 0 | 7.44 | 100 | **0** | 5.69 | 100 | 0 | **5.06** |
| | $\alpha=0$ | 72 | 13.67 | 22.38 | 78 | 11.63 | 21.81 | 75 | 11.41 | 20.81 | 75.5 | 11.55 | 20.25 | 76 | 11.45 | 20.31 |
| | $\alpha=1$ | 99.5 | 0.99 | 18.5 | 100 | 0.04 | 14.25 | 100 | 0 | 11.5 | 100 | 0 | 8.69 | 100 | 0 | 7.19 |

*Table 4  Results with Test 4, different databases and VQ-based automatic signature verifier $\alpha_{OPT}$ denotes the optimal alpha value for considering both set of features at score level. $\alpha = 0$ and $\alpha = 1$ mean that only the second or first set of features, respectively, was evaluated. Green color highlights the optimal VQ configuration for random forgeries. Yellow color highlights the optimal VQ configuration for skilled forgeries. This is always obtained for codebook sizes greater or equal to six bits.*

| | | b=4 | | | b=5 | | | b=6 | | | b=7 | | | b=8 | | |
|---|---|---|---|---|---|---|---|---|---|---|---|---|---|---|---|---|
| | | IDR | DCFr | DCFs | IDR | DCFr | DCFs | IDR | DCFr | DCFs | IDR | DCFr | DCFs | IDR | DCFr | DCFs |
| MCYT VQ1_100 | $\alpha_{OPT}$ | 97 | 3.38 | 12.72 | 99 | 2.22 | 9.52 | 98.8 | 1.42 | 7.86 | 98.8 | 1.92 | 7.3 | 98 | 2.47 | 6.98 |
| | $\alpha=0$ | 17.6 | 26.37 | 29.44 | 20.2 | 26.20 | 29.34 | 20.8 | 26.77 | 28.9 | 18.6 | 28.11 | 29.6 | 16.6 | 30.53 | 31.18 |
| | $\alpha=1$ | 96.6 | 3.91 | 13.22 | 98.8 | 2.45 | 9.76 | 98.6 | 1.44 | 8.12 | 98.8 | 1.92 | 7.5 | 98 | 2.49 | 7.08 |
| MCYT VQ5_100 | $\alpha_{OPT}$ | 100 | 1.97 | 10.42 | 100 | 0.69 | 7.06 | 100 | 0.54 | 4.96 | 100 | 0.35 | 3.92 | 100 | 0.34 | 3.28 |
| | $\alpha=0$ | 39.6 | 19.16 | 24.02 | 48 | 17.19 | 22.72 | 50 | 17.06 | 22.26 | 50 | 15.97 | 20.96 | 51.6 | 15.96 | 20.84 |
| | $\alpha=1$ | 100 | 2.65 | 12.04 | 100 | 1.01 | 7.8 | 100 | 0.66 | 5.38 | 100 | 0.38 | 4.46 | 100 | 0.40 | 3.76 |
| MCYT VQ1_330 | $\alpha_{OPT}$ | 92.24 | 4.20 | 13.10 | 96 | 2.94 | 10.33 | 96.48 | 2.38 | 8.94 | 96.67 | 2.48 | 8.30 | 96 | 2.75 | 8.18 |
| | $\alpha=0$ | 14.91 | 27.12 | 31.72 | 16.24 | 27.02 | 31.38 | 15.21 | 26.93 | 30.86 | 14.85 | 27.68 | 31.15 | 13.09 | 29.70 | 32.24 |
| | $\alpha=1$ | 92.12 | 4.22 | 13.44 | 96 | 2.97 | 10.50 | 96.42 | 2.39 | 9.08 | 96.48 | 2.48 | 8.36 | 96 | 2.75 | 8.18 |
| MCYT VQ5_330 | $\alpha_{OPT}$ | 99.09 | 2.14 | 9.72 | 99.39 | 1.00 | 6.46 | 99.39 | 0.78 | 4.79 | 99.39 | 0.70 | 4.06 | 99.39 | 0.67 | 3.58 |
| | $\alpha=0$ | 31.09 | 18.60 | 25.44 | 36.79 | 17.27 | 24.23 | 39.51 | 16.86 | 23.90 | 42.06 | 16.55 | 23.60 | 42.30 | 16.02 | 23.58 |
| | $\alpha=1$ | 99.09 | 2.33 | 10.51 | 99.39 | 1.09 | 6.94 | 99.39 | 0.79 | 4.95 | 99.39 | 0.72 | 4.15 | 99.39 | 0.67 | 3.58 |
| SVC VQ1 | $\alpha_{OPT}$ | 99.5 | 1.51 | 19.13 | 99.5 | 0.84 | 17 | 99.5 | 0.78 | 14.13 | 99.5 | 1.29 | 13.31 | 99.5 | 1.56 | 12.94 |
| | $\alpha=0$ | 60 | 18.01 | 30.13 | 62.5 | 16.98 | 28.38 | 63 | 15.99 | 27 | 61 | 16.81 | 27.75 | 53 | 18.47 | 28.38 |
| | $\alpha=1$ | 99.5 | 1.63 | 22 | 99.5 | 0.94 | 19.31 | 99.5 | 0.78 | 17.63 | 99.5 | 1.29 | 15.88 | 99.5 | 1.56 | 14.94 |
| SVC VQ5 | $\alpha_{OPT}$ | 100 | 0.33 | 12.56 | 100 | 0.01 | 9.88 | 100 | 0 | 7.69 | 100 | 0 | 6.75 | 100 | 0 | 5.44 |
| | $\alpha=0$ | 90 | 8.20 | 20.75 | 89.5 | 7.75 | 17.38 | 93 | 6.57 | 16.06 | 92.5 | 6.46 | 15.31 | 91.5 | 6.30 | 15.19 |
| | $\alpha=1$ | 99.5 | 0.78 | 19.19 | 100 | 0.04 | 14.13 | 100 | 0 | 10.75 | 100 | 0 | 8.75 | 100 | 0 | 7.63 |

*Table 5  Results with all tests, different databases and DTW-based automatic signature verifier $\alpha_{OPT}$ denotes the optimal alpha value for considering both set of features at score level. $\alpha = 0$ and $\alpha = 1$ mean that only the second or first set of features, respectively, was evaluated. Green color highlights the optimal results for DTW and random forgeries. Yellow color highlights the optimal results for DTW and skilled forgeries. In contrast to VQ, there is no model size parameter to adjust in this situation.*

| | MCYT100 | | | | | | MCYT330 | | | | | | SVC | | | | | |
|---|---|---|---|---|---|---|---|---|---|---|---|---|---|---|---|---|---|---|
| | DTW1 | | | DTW5 | | | DTW1 | | | DTW5 | | | DTW1 | | | DTW5 | | |
| | IDR | DCFr | DCFs | IDR | DCFr | DCFs | IDR | DCFr | DCFs | IDR | DCFr | DCFs | IDR | DCFr | DCFs | IDR | DCFr | DCFs |
| TEST1$\alpha_{OPT}$ | 99.4 | 0.55 | 7.87 | 99.4 | 0.56 | 3.48 | 98 | 1.30 | 8.64 | 99.21 | 0.59 | 3.66 | 100 | 0.2 | 24.79 | 100 | 0.18 | 10.10 |
| TEST1$\alpha=0$ | 68 | 12.39 | 24.94 | 84 | 6.47 | 10.2 | 55.21 | 14.09 | 26.43 | 76.79 | 7.26 | 9.72 | 48.5 | 18.83 | 34 | 89 | 10.91 | 14.10 |
| TEST1$\alpha=1$ | 99 | 0.66 | 8.51 | 99.4 | 0.66 | 3.58 | 97.70 | 1.35 | 9.26 | 99.21 | 0.62 | 3.86 | 100 | 0.2 | 26.31 | 100 | 0.22 | 11.83 |
| TEST2$\alpha_{OPT}$ | 99.6 | 0.55 | 7.57 | 99.4 | 0.64 | 3.48 | 97.94 | 1.26 | 8.69 | 99.21 | 0.62 | 3.59 | 100 | 0.13 | 21.10 | 100 | 0.17 | 10.35 |
| TEST2$\alpha=0$ | 41 | 22.65 | 32.79 | 61.8 | 18.92 | 9.55 | 31.76 | 23.79 | 33.96 | 42 | 18.88 | 9.69 | 82.5 | 9.90 | 28.04 | 96.5 | 8.53 | 14.56 |
| TEST2$\alpha=1$ | 99 | 0.66 | 8.51 | 99.4 | 0.66 | 3.58 | 97.70 | 1.35 | 9.26 | 99.21 | 0.62 | 3.86 | 100 | 0.20 | 26.31 | 100 | 0.22 | 11.83 |
| TEST3$\alpha_{OPT}$ | 100 | 0.38 | 7.32 | 99.6 | 0.58 | 3.18 | 98.73 | 0.94 | 8.14 | 99.45 | 0.57 | 3.31 | 100 | 0.13 | 18.98 | 100 | 0.17 | 9 |
| TEST3$\alpha=0$ | 15.2 | 34.37 | 40.86 | 33.4 | 26.33 | 14.25 | 9.39 | 34.65 | 41.44 | 20 | 26.30 | 13.60 | 66 | 16.29 | 33.92 | 85 | 13.10 | 18.88 |
| TEST3$\alpha=1$ | 100 | 0.38 | 7.56 | 99.6 | 0.58 | 3.51 | 98.48 | 0.95 | 8.29 | 99.39 | 0.57 | 3.50 | 100 | 0.17 | 21.67 | 100 | 0.17 | 10.04 |
| TEST4$\alpha_{OPT}$ | 100 | 0.35 | 7.22 | 99.6 | 0.58 | 3.41 | 98.79 | 0.91 | 8.21 | 99.45 | 0.57 | 3.35 | 100 | 0.14 | 18.58 | 100 | 0.17 | 9.29 |
| TEST4$\alpha=0$ | 41 | 22.65 | 32.79 | 61.8 | 18.92 | 9.55 | 31.76 | 23.79 | 33.96 | 42 | 18.88 | 9.69 | 82.5 | 9.90 | 28.04 | 96.5 | 8.53 | 14.56 |
| TEST4$\alpha=1$ | 100 | 0.38 | 7.56 | 99.6 | 0.58 | 3.51 | 98.48 | 0.95 | 8.29 | 99.39 | 0.57 | 3.50 | 100 | 0.17 | 21.67 | 100 | 0.17 | 10.04 |

TEST1

Overall, we found that pressure improved the verification rates when combined with other modalities. The higher combined ratios demonstrated this for $\alpha \neq 1$ compared to the case where the pressure was not used. In particular, we observed that the pressure effectively reduced the error rates for skilled and random forgeries in most database.

For example, in MCYT100 VQ1, we found that the pressure consistently reduced the performance, with fewer bits resulting in higher contributions, as shown in Table 1. In addition, we found that 7 bits produced the best performance for both random and skilled forgeries. Similarly, in MCYT100 VQ5, the pressure improved the results by roughly 1.5% and 0.5% for skilled and random forgeries, respectively. The best results were achieved with 7 and 8 bits.

In MCYT330 VQ1 and VQ5, we observed that the fusion constantly improved the results, with the random forgery being less reduced than the skilled forgery, as expected. The best results were found with 6 and 7 bits for random and skilled forgeries, respectively, in VQ1 and 8 bits in VQ5. Moreover, we found that the higher the number of training/reference/enrolled signatures, the better the results obtained with higher bits.

However, in SVC2004 VQ1 and VQ5, we found that add-ing pressure did not improve performance. In fact, adding pressure degraded the performance in VQ1, while in VQ5, no contribution was observed. After fusion, the best performance was obtained with $\alpha = 1$ for skilled forgeries and $\alpha = 0.96$ for random forgeries, with 8 and 7 bits, respectively. In MCYT100 DTW1 and DTW5 (Table 5), we found that the fusion improved the results in all cases, but adding pressure did not contribute significantly. The contribution was 0.1% for random forgeries and 0.6% for skilled forgeries, which reported higher performance. Similarly, in MCYT-330 DTW1 and DTW5, we observed almost the same performance when $(x, y)$ were only used to verify signatures in random forgery and a slightly improved performance (less than 1% of error) in skilled forgery.

In SVC2004 DTW1 and DTW5, we observed a reduction of about 2% in error rates for skilled forgeries when pressure was added. This high reduction was due to the high error rates observed. We also found that the algorithm selected 0.5 as the optimal $\alpha$, which means that both sets of features have the same importance. In general, we conclude that the hardware used to acquire pressure provides a signal with low discriminative potential in signature verification. However, when combined with other modalities, a modest improvement in verification rates can be consistently observed across the databases and experimental protocols.

TEST2

Based on Tables 2 and 5, the results of TEST2 show that adding angles to the pressure in the same set can improve the performance in some cases for both automatic signature verification (ASV) tasks but can also lead to degradation in some cases. In MCYT100 VQ1, adding angles improved the random forgery slightly but degraded the skilled forgeries results compared to TEST1. In MCYT100 VQ5, both random and skilled forgeries experiments were somewhat improved. In practical terms, adding or not adding angles produced the same performance after fusion.

For MCYT330 VQ5, the performance was slightly degraded and slightly improved in random and skilled for-geries. However, no effects were observed by adding the angles along the pressure in the same set. In SVC2004 VQ1, the performance was reduced by about 2.5% for skilled forgeries compared to TEST1, with a stable effect in random forgeries. In SVC2004 VQ5, better improvements were observed in skilled forgeries compared to TEST1 (reduction of 3% of error) and compared to the use of only $y$ (reduction of 3.5% of error).

No effects were observed for MCYT100 DTW1 and DTW5 compared to TEST1, except for a slightly negative result in random forgery in MCYT100 DTW5. Similarly, no effects were observed in MCYT330 DTW1 and DTW5, except for a somewhat negative impact in random forgery in MCYT330 DTW5. In SVC2004 DTW1, an improvement in skilled forgeries was observed compared to TEST1, prob-ably due to the high minDCF error. A non-important performance advance was observed in random forgery. However, SVC2004 DTW5 showed stable results compared to TEST1.

In general, the results from TEST2 suggest that adding angles to the pressure in the same set can improve the performance in some cases for both ASVs. In contrast, in other cases, the performance is stable, and in rare cases, it can be slightly degraded.

TEST3

The results of MCYT100 VQ1 in Table 3 show that no significant effect was observed in the fusion due to the values of α being close to 1. However, the performance in both random and skilled forgeries improved slightly after adding angles along the pressure in the same set.

MCYT100 VQ5 showed that adding a set with angles did not affect random forgeries and a small positive effect on skilled forgeries. MCYT330 VQ1 and VQ5 produced similar results, with no significant effects observed when angles were added in a new set.

SVC2004 VQ1 showed that adding angles along with the pressure degraded the performance of random forgery compared to TEST1 and TEST2, but there were no effects observed with skilled forgeries. In contrast, SVC2004 VQ5 did not show any improvements in random forgery, but it considerably improved the performance of skilled forgery by approximately 2% of error.

In Table 5, we observed that MCYT100 DTW1 and DTW5 showed no effects on random forgery but a modest improvement in skilled forgery performance. Similar findings were observed in MCYT330 DTW1 and DTW5, where adding angles did not show any improvement in random forgery, but there was a slight reduction in skilled forgery errors by 0.2%. SVC2004 DTW1 showed no effects on random forgery, but there was a relevant positive effect on skilled forgery performance. In contrast, SVC2004 DTW5 showed no impact on random forgery, but it positively affected skilled forgery with a reduction of 1% of error.

In general, the findings from TEST3 cannot be directly compared to TEST1 or TEST2 because the first set of fea- tures was changed. However, the results suggest that adding pressure to the same features as $x$ and $y$ can improve performance. Adding angles along with the pressure can positively impact performance, but the impact is smaller for better classifiers and varies across the datasets. Overall, TEST3 produced the best performance among all the tests. The results also suggest that the impact of adding angles is positive in some cases and does not produce any adverse effects in most cases, except for rare cases.

TEST4

The performance of MCYT100 VQ1 in Table 4 was observed to be the same as that of TEST3, with no relevant effect found in the fusion. However, the performance in random and skilled forgeries improved slightly in MCYT100 VQ5, with a reduction of 0.1% and 0.5%, respectively, in both experiments.
No effects were found when fusing set 2 in MCYT330 VQ1, and a similar finding was observed in MCYT330 VQ5. The more reference signatures and users in the data- base, the less impact there was on the performance by fusing set 2. In contrast, fusing set 2 produced good improvements in skilled and zero effects in random forgeries in SVC2004 VQ1 and SVC2004 VQ5.

In MCYT100 DTW1 in Table 5, both random and skilled forgery performances were somewhat reduced. Observations to MCYT100 DTW5 showed that only the skilled forgery was reduced (0.1% of error), with no effect on random forgery. Alphas were around 0.9 in MCYT330 DTW1, resulting in modest effects in both experiments. MCYT330 DTW5 showed a slightly positive impact in skilled forgeries.

SVC2004 DTW1 showed an almost stable effect in random forgeries but a high reduction in skilled forgeries (from minDCF = 21.67% to minDCF = 18.58%). The same observation was made in SVC2004 DTW5, with a modest performance reduction in skilled forgeries. Skilled forgeries detection was found to be the most challenging experiment. In some cases, performance improved. However, improvements were not observed in the case of random forgeries, possibly because the performance in random forgeries was quite low compared to skilled forgeries.
Comparing combined 4 with combined 3, an improvement was observed in set 4 for VQ5_100, but not signifi- cantly in DTW. Replication of $p$ in the second set did not produce a significant improvement. In general, the study showed that while the impact of different techniques varied depending on the experiment, the global best performance was observed in TEST3.

**Conclusions**

In summary, this paper has investigated the impact of common features in signature verification using two automatic signature verifiers with various training signatures and databases. Our analysis focused on five features acquired by the majority of digitizers, namely the signature trajectory $(x, y)$, pressure ($p$), and pen-tip angles (azimuth and elevation). We observed that the combination of pressure with and $y$ features resulted in a consistent improvement in performance across the databases and experimental protocols. However, the pressure alone had low discriminative potential in signature verification. Adding the pen-tip angles to the pressure in the same feature set produced mixed results, with some cases showing improved performance and others remaining stable or slightly degraded. Moreover, we observed that including pressure in the same set of features as and $y$ can lead to better performance. Overall, our results suggest that careful selection and com- bination of signature features are crucial for improving the accuracy of automatic signature verification systems. Worth to mention that this feature vector splitting can be applied to other classifiers to reduce the dimensional- ity problem and to bring light to the contribution of each feature to the classification accuracies. While we have applied the vector splitting strategy in an e-security application (biometric recognition of people), it can also be used in e-health applications.

Conference Eurospeech, vol. 4. 1997. p. 1895–8.